\documentclass[10pt,twocolumn,letterpaper]{article}
\usepackage{cvpr}
\usepackage{epsfig}
\usepackage{times}
\usepackage[utf8]{inputenc} 
\usepackage[T1]{fontenc}    
\usepackage{url}            
\usepackage{booktabs}       
\usepackage{amsfonts}       
\usepackage{nicefrac}       
\usepackage{microtype}      
\usepackage{lipsum} 
\usepackage{color}
\usepackage{graphicx}
\usepackage{subcaption}
\usepackage{xspace}
\usepackage{booktabs} 
\usepackage{amsmath}
\usepackage{amssymb}
\usepackage{multirow}
\usepackage{multicol}
\usepackage{algorithm}
\usepackage{algpseudocode}
\usepackage{adjustbox}
\usepackage{flushend}

\usepackage[pagebackref=true,breaklinks=true,letterpaper=true,colorlinks,bookmarks=false]{hyperref}

\cvprfinalcopy 

\def\cvprPaperID{440} 

\ifcvprfinal\fi

\newcommand{\model}{TAFE-Net\xspace}
\newcommand{\modelplural}{TAFE-Nets\xspace}

\renewcommand{\vec}[1]{\mathbf{#1}}
\newcommand\minisection[1]{\vspace{2mm}\noindent \textbf{#1}}

\title{TAFE-Net: Task-Aware Feature Embeddings for Low Shot Learning}

\author{
	Xin Wang\quad Fisher Yu\quad 
	Ruth Wang\quad 
	Trevor Darrell\quad Joseph E. Gonzalez \\
	{UC Berkeley}
}

\begin{document}
	
	\maketitle
	
	\begin{abstract}
		Learning good feature embeddings for images often requires substantial training data.
		As a consequence, in settings where training data is limited (e.g., few-shot and zero-shot learning), we are typically forced to use a generic feature embedding across various
		 tasks. Ideally, we want to construct feature embeddings that are tuned for the given task.
		In this work, we propose Task-Aware Feature Embedding Networks (\modelplural\footnote{Pronounced taffy-nets}) to learn how to adapt the image representation to a new task in a meta learning fashion. 
		Our network is composed of a meta learner and a prediction network. Based on a task input, the meta learner generates parameters for the feature layers in the prediction network so that the feature embedding can be accurately adjusted
		for that task.
		We show that \model is highly effective in generalizing to new tasks or concepts and evaluate the \model
		on a range of benchmarks in zero-shot and few-shot learning.
		Our model matches or exceeds the state-of-the-art on all tasks. In particular, our approach improves the
		prediction accuracy of unseen attribute-object pairs by 4 to 15 points on the challenging visual 
		attribute-object composition task.
	\end{abstract}
	
	\section{Introduction}
Feature embeddings are central to 
computer vision.
By mapping images into semantically rich vector spaces, feature embeddings extract key information that can be used for a wide range of prediction tasks.  
However, learning good feature embeddings typically requires substantial amounts of training data and computation.  
As a consequence, a common practice~\cite{donahue2014decaf,girshick2014rich,zeiler2014visualizing} is to re-use existing feature embeddings from convolutional networks (e.g., ResNet~\cite{he2016deep}, VGG~\cite{simonyan2014very}) trained on large-scale labeled training datasets (e.g., ImageNet~\cite{russakovsky2015imagenet}); to achieve maximum accuracy, these generic feature embedding are often fine-tuned~\cite{donahue2014decaf,girshick2014rich,zeiler2014visualizing} or transformed~\cite{Hoffman_cycada2017} using additional task specific training data. 

\begin{figure}[t]
	\centering
	\includegraphics[width=0.48\textwidth]{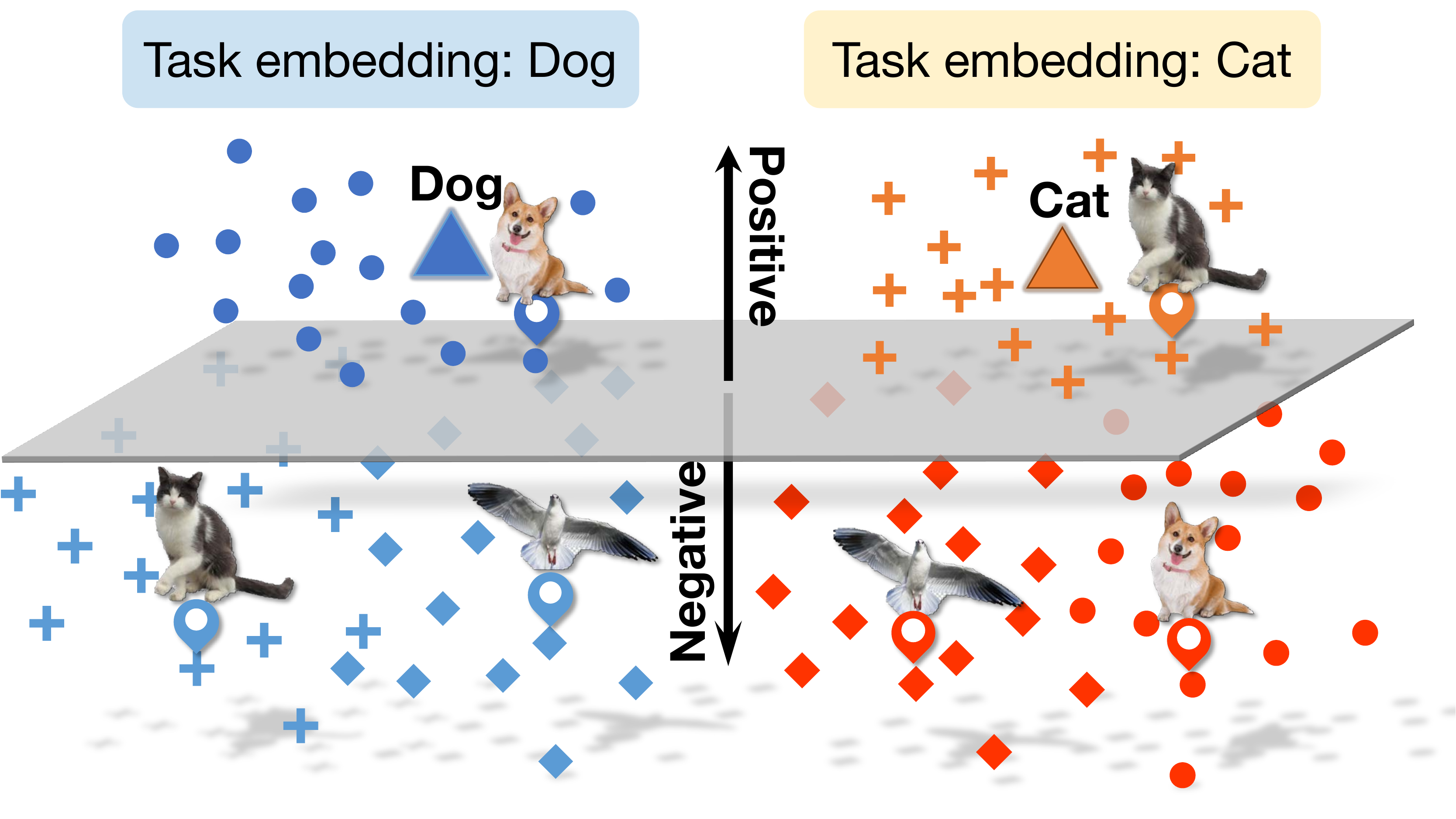} \caption{A cartoon illustration of Task-aware Feature Embeddings (TAFEs). In this case there are two binary prediction tasks: \texttt{hasCat} and \texttt{hasDog}.
		Task-aware feature embeddings mean that the same image can have different embeddings for each task.
		As a consequence, we can adopt a single task independent classification boundary for all tasks.\vspace{-1em}}
	\label{fig:tafe}
\end{figure}

In many settings, the training data are insufficient to learn or even adapt generic feature embeddings to a given task.
For example, in zero-shot and few-shot prediction tasks, the scarcity of training data forces the use of generic feature embeddings~\cite{lampert2014attribute,xian2018zero,zhang2016zero}.
As a consequence, in these situations, much of the research instead focuses on the design of joint task and data embeddings~\cite{Changpinyo_2016_CVPR,frome2013devise,zhang2016zero} that can be generalized to unseen tasks or tasks with fewer examples. 
Some have proposed treating the task embedding as linear separators and learning to generate them for new tasks~\cite{vinyals2016matching,lu2016visual}. 
Others have proposed hallucinating additional training data~\cite{xian2018feature,hariharan2017low, wang2018low}. 
However, in all cases, a common image embedding is shared across tasks. Therefore, the common image embedding may be out of the domain or sub-optimal for any individual prediction task and may be even worse for completely 
new tasks. This problem is exacerbated in settings where the number and diversity of training tasks is relatively small~\cite{finn2017one}.

\begin{figure*}[t]
	\centering
	\includegraphics[width=1\textwidth]{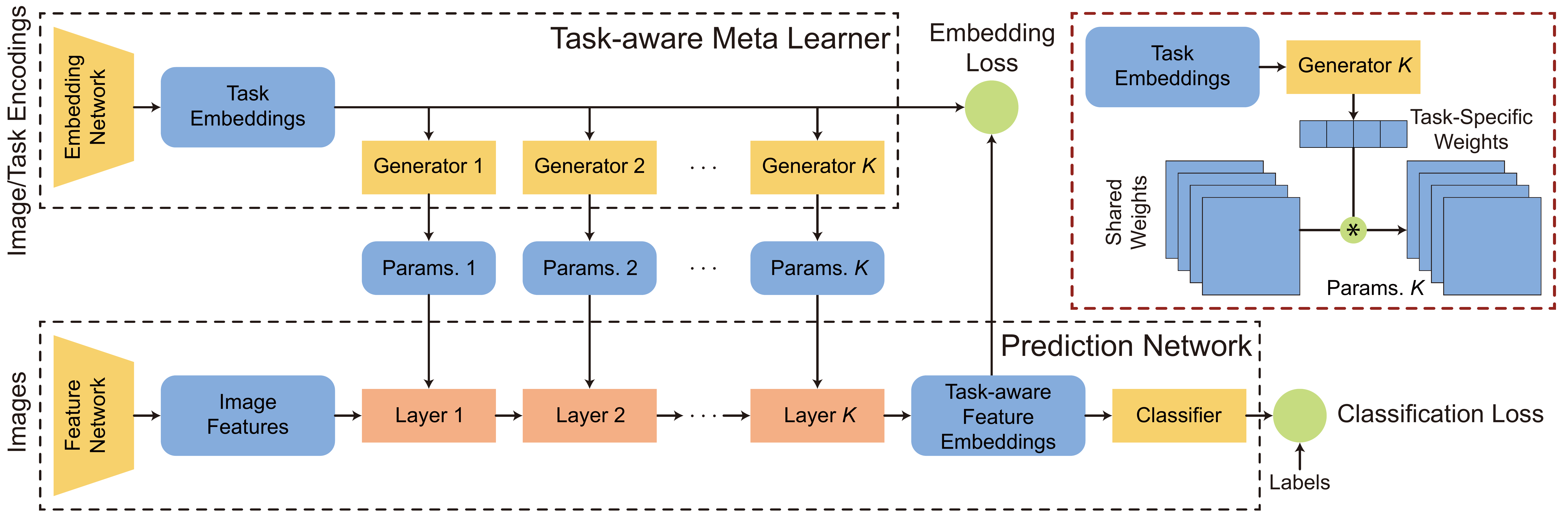}
	\caption{\model architecture design. \model has
		a task-aware meta learner that generates the parameters of
		the feature layers within the classification subnetwork to transform the generic image features to TAFEs. The generated weights are
		factorized into low-dimensional task-specific weights and 
		high-dimensional shared weights across all tasks to reduce the complexity of the parameter generation. A single 
		classifier is shared across all tasks taking the resulting TAFEs as inputs.\vspace{-1em}}
	\label{fig:model_arch}
\end{figure*}

In this work, we explore the idea of dynamic feature representation by introducing the task-aware feature embedding network (\model) with a meta-learning based 
parameter generator to transform generic image features to task-aware feature embeddings (TAFEs). As illustrated in Figure~\ref{fig:tafe}, the representation of TAFEs is adaptive to the given semantic task description, and thus able to accommodate the need of new tasks at testing time. The feature transformation is realized with a 
task-aware meta learner, which generates the parameters of feature embedding layers within the classification subnetwork shown in Figure~\ref{fig:model_arch}.  Through the use of TAFEs, we can adopt a simple binary classifier to learn a task-independent linear boundary that can separate the positive and negative examples and generalize to new tasks. 

We further propose two design innovations to address the challenges due to the limited number of training tasks~\cite{finn2017one} and the complexity of the parameter generation~\cite{bertinetto2016learning}. 
Dealing with the limited tasks, we couple the
task embedding to the task aware feature embeddings with a novel embedding loss based on metric
learning. The resulting coupling improves generalization across tasks by jointly clustering both images and
tasks. Moreover, the parameter generation requires predicting a large number of weights from a low dimensional task embedding (e.g., a 300-dimensional vector extracted with GloVe~\cite{pennington2014glove}), which can be 
complicated and even infeasible to train in practice, we therefore introduce a novel decomposition to factorize
the weights into a small set of task-specific weights needed for generation on the fly and a large set of static
weights shared across all tasks.

We conduct an extensive experimental evaluation in Section~\ref{sec:exp}. The proposed \model exceeds the state-of-the-art zero-shot learning approaches on three out of five standard benchmarks (Section~\ref{sec:zeroshot}) without the need of additional data generation, a complementary approach that has shown boosted performance compared to mere discriminative models by the recent work~\cite{xian2018feature}. On the newly proposed unseen attribute-object composition recognition task~\cite{misra2017red}, we are able to achieve an improvement of 4 to 15 points over the state-of-the-art (Section~\ref{sec:concept}). Furthermore, the proposed
architecture can be naturally applied to few-shot learning (Section~\ref{sec:fewshot}), achieving competitive results on the ImageNet based
benchmark introduced by Hariharan \etal~\cite{hariharan2017low}. 
The code is available at \href{https://github.com/ucbdrive/tafe-net}{https://github.com/ucbdrive/tafe-net}.

	\section{Related Work}
Our work is related to several lines of research in zero-shot learning as well as parameter generation, dynamic neural 
network designs, and feature modulation. Built on top of the rich prior works, to the best of our knowledge, we are the first to study dynamic image feature representation for zero-shot and few-shot learning. 

Zero-shot learning falls into the multimodal learning regime 
which requires a proper leverage of multiple sources (e.g., image features and semantic embeddings of the tasks). Many~\cite{koch2015siamese, yang2018learning, vinyals2016matching, zhang2016zero, Changpinyo_2016_CVPR, frome2013devise} 
have studied metric learning based objectives to jointly learn the task embeddings and image embeddings, resulting in a similarity
or compatibility score that can later be used for classification~\cite{misra2017red,vinyals2016matching,lampert2014attribute,akata2016label,akata2015evaluation,frome2013devise,socher2013zero}. Conceptually, our
approach shares the \emph{matching} spirit with the introduction of a 
binary classifier which predicts whether or not the input image matches the task description. 
In contrast to prior works, we transform the image features according to the task and thus we only need to learn a task-independent 
decision boundary to separate the positive and negative examples similar to the classic supervised learning. The proposed embedding loss in our work also adopts metric learning for joint embedding 
learning but with the main goal to address the limited number of training tasks in meta learning~\cite{finn2017one}. More 
recently, data hallucination has been used in the zero-shot~\cite{xian2018feature,zhu2018generative} and few-shot~\cite{hariharan2017low,wang2018low} learning which indicate that the additional synthetic data of the unseen tasks are useful to learn the classifier and can be augmented with the discriminative models~\cite{xian2018feature,wang2018low}. Our (discriminative) model does not utilize additional data points and we show in experiments that our model can match or outperform the 
generative models on a wide range of benchmarks. We believe the approaches requiring additional data generation can benefit from a 
stronger base discriminative model.

\model uses a task-aware meta learner to generate parameters of the
feature layers. Several efforts~\cite{bertinetto2016learning,ha2016hypernetworks,denil2013predicting} have studied the idea of
adopting one  meta network to generate weights of another network.
Our task-aware meta learner serves a similar role for the weight generation 
but in a more structured and constrained manner. We study different mechanisms to decompose the weights of the prediction network so that it can generate weights for multiple layers at once. In contrast, Bertinetton~\etal~\cite{bertinetto2016learning} 
focus on generating weights for a single layer and Denil~\etal~\cite{denil2013predicting} can generate only up to 95\% parameters of a single layer due to the quadratic size of the output space. 


The \model design is also related to works on dynamic neural networks~\cite{wang2018skipnet,wu2018blockdrop,wang2017idk,lin2017runtime} which focus on dynamic execution at runtime. SkipNet~\cite{wang2018skipnet} proposed by Wang~\etal introduces recurrent gating to dynamically control the network activations based on the input. In contrast, \model 
dynamically re-configures the network parameters rather than the network structure as in the prior works~\cite{wang2018skipnet,wu2018blockdrop} aiming to learn adaptive image features for the given task. 

In the domain of visual question answering, previous works~\cite{perez2018film, de2017modulating} explore the use of a question embedding network to modulate the features of the primary convolutional network. Our factorized weight
generation scheme for convolutional layers can also be viewed as channel-wise feature modulation. However, the proposed parameter generation framework is more general than feature modulation which can host different factorization strategies~\cite{bertinetto2016learning}.


	\section{Task-Aware Feature Embedding}
As already widely recognized, feature embeddings are the fundamental building blocks for many applications~\cite{krizhevsky2012imagenet,long2015fully,girshick2015fast} in 
computer vision. In this work, we introduce task-aware feature embeddings (TAFEs), a type of dynamic image feature representation that adapts to the given task. We demonstrate that such dynamic feature representation has applications in the zero-shot learning, few-shot learning and unseen attribute-object pair recognition.

We start with the \model model design in Section~\ref{sec:model} and then introduce the weight factorization (Section~\ref{sec:weight_factorization}) and the embedding loss (Section~\ref{sec:embed_loss}) to address the
challenges with the weight generation and the limited number of training tasks. We delay the specifications of 
different task descriptions and the setup of various applications to Section~\ref{sec:app}. 

\subsection{\model Model}
\label{sec:model}
There are two sub-networks
in \model as shown in Figure~\ref{fig:model_arch}: a 
task-aware meta leaner $\mathcal{G}$ and a prediction network
$\mathcal{F}$. The task-aware meta learner takes a task description $\Vec{t}\in\mathcal{T}$ (e.g., word2vec~\cite{mikolov2013efficient} encoding or example images, detailed in Section~\ref{sec:app}) and generates the weights of the feature layers in the prediction network. 

For an input image $\Vec{x}\in\mathcal{X}$, the prediction network:
\begin{equation}
\mathcal{F}(\Vec{x};\theta_t) = \Vec{y},    
\end{equation}
predicts a binary label $\Vec{y}\in\mathcal{Y}$ indicating whether or not the input image $x$ is compatible with 
the task description $\Vec{t}$. More specifically, we adopt a pre-trained feature extractor on ImageNet (e.g., ResNet~\cite{he2016deep}, VGG~\cite{simonyan2014very} whose parameters are frozen during training) to produce generic features of the input images and then feed the generic features to a sequence of \emph{dynamic}
feature layers whose parameters denoted by $\theta_t$ are generated by $\mathcal{G}(\Vec{t})$.  The output of the dynamic feature layers is named 
as \emph{task-aware feature embedding} (TAFE) in the sense that the feature embedding 
of the same image can be different under different task descriptions. Though not directly used as the input to $\mathcal{F}$,
the task description $\Vec{t}$ controls the parameters of the feature layers in $\mathcal{F}$ and further injects the task information to the
image feature embeddings. 

We are now able to introduce a simple binary classifier in $\mathcal{F}$, which takes TAFEs as inputs,  
to learn a task-independent decision boundary. When multi-class predictions are needed, we can leverage the predictions of $\mathcal{F}(\Vec{x})$ under different tasks descriptions and use them as probability scores. The objective formulation is presented in Section~\ref{sec:embed_loss}. 

The task-aware meta learner $\mathcal{G}$ paramterized by $\eta$ is composed of an embedding network 
$\mathcal{T}(\Vec{t})$ to generate a task embedding $\mathbf{e}_t$ and a set of
weight generators $\mathbf{g}^i, i=\{1...K\}$ that generate parameters for K dynamic feature
layers in $\mathcal{F}$ conditioned on the same task embedding $\mathbf{e}_t$.  

\subsection{Weight Generation via Factorization}
\label{sec:weight_factorization}
We now present the weight generation scheme for the feature layers in 
$\mathcal{F}$. The feature layers that produce the task aware feature embeddings (TAFE) can either be convolutional layers or fully-connected (FC) layers. To generate the feature layer weights,  we will need the output dimension of $\mathbf{g}^i$ (usually a FC layer) to match the weight size of the $i$-th feature layer in $\mathcal{F}$. As noted by Bertinetto~\etal~\cite{bertinetto2016learning}, the 
number of weights required for the meta-learner estimation is often much greater than that of the task descriptions Therefore, it is difficult to learn weight generation from a small number of example tasks. Moreover, the parametrization of the weight generators $\mathbf{g}$ can consume a large amount 
of memory, which makes the training costly and even infeasible. 

To make our meta learner generalize effectively, we propose a weight 
factorization scheme along the output dimension of each FC layer and 
the output channel dimension of a convolutional layer.
This is distinct from the low-rank decomposition used in prior meta-learning works~\cite{bertinetto2016learning}. 
The channel-wise factorization builds on the intuition 
that channels of a convolutional layer
may have different or even orthogonal functionality. 

\minisection{Weight factorization for convolutions}. 
Given an input tensor $\Vec{x}^i\in \mathbb{R}^{w\times h\times c_\text{in}}$ for
the $i\text{-th}$ feature layer in $\mathcal{F}$ whose weight is $\Vec{W}^i \in\mathbb{R}^{k\times k\times c_\text{in}\times c_\text{out}}$ 
($k$ is the filter support size and $c_\text{in}$ and $c_\text{out}$ are 
the number of input and output channels) and bias is  $\Vec{b}^i\in\mathbb{R}^{c_\text{out}}$, the output $\Vec{x}^{i+1}\in \mathbb{R}^{w'\times h'\times c_\text{out}}$ of the convolutional layer is given
by 
\begin{equation}
\Vec{x}^{i+1} = \Vec{W}^i * \Vec{x}^{i} + \Vec{b}^{i},
\end{equation}
where $*$ denotes convolution. Without loss of generality, we remove the bias term of the convolutional layer as it is often followed by batch normalization~\cite{ioffe2015batch}. $\Vec{W}^i = \Vec{g}^i(\Vec{t})$ is
the output of the $i\text{-th}$ weight generator in $\mathcal{G}$ in the full weight generation setting.  
We now decompose the weight $\Vec{W}^i$ into 
\begin{equation}
\Vec{W}^i=\Vec{W}^i_s *_{c_{\text{out}}} \Vec{W}^i_t,
\end{equation}
where $\Vec{W}^i_s\in\mathbb{R}^{k\times k \times c_\text{in}\times c_\text{out}}$ is a shared parameter aggregating all tasks $\{\Vec{t}_1, ...\Vec{t}_T\}$ and $\Vec{W}_t\in\mathbb{R}^{1\times1\times c_\text{out}}$
is a task-specific parameter depending on the current task input. $*_{c_\text{out}}$ denotes the grouped convolution along the output channel dimension,
i.e. each channel of $x*_{c_\text{out}} y$ is simply the convolution of the corresponding channels in $x$ and $y$. 
The parameter generator $\Vec{g}^i$ only needs to generate 
$\Vec{W}^i_t$ which reduces the output dimension of $\Vec{g}^i$ from $k\times k \times c_\text{in}\times c_\text{out}$ to $c_\text{out}$.

\minisection{Weight factorization for FCs}. 
Similar to the factorization of the convolution weights,
the FC layer weights $\Vec{W}^i\in\mathbb{R}^{m\times n}$ can be decomposed into 
\begin{equation}
\Vec{W}^i = \Vec{W}^i_s \cdot \text{diag}(\Vec{W}^i_t),
\end{equation}
where $ \Vec{W}^i_s\in\mathbb{R}^{m\times n}$ is the shared parameters for
all tasks and $\Vec{W}^i_t\in\mathbb{R}^{n}$ is the task-specific parameter. Note that this factorization 
is equivalent to the feature activation modulation, that is, for an input $\Vec{x}\in\mathbb{R}^{1\times m}$, 
\begin{equation}
\vec{x}\cdot(\Vec{W}^i_s \cdot \text{diag}(\Vec{W}^i_t))= (\Vec{x}\cdot\Vec{W}^i_s) \odot \Vec{W}^i_t,
\end{equation}
where $\odot$ denotes element-wise multiplication. 

As a consequence,  the weight generators only need to generate low-dimensional
task-specific parameters for each task in lower dimension and learn one set of high 
dimensional parameters shared across all tasks. 

\subsection{Embedding Loss for Meta Learner}
\label{sec:embed_loss}
The number of task descriptions used for training the
task-aware meta learner is usually much smaller than the number of images available
for training the prediction network. The data scarcity issue may lead to a 
degenerate meta learner. We, therefore, propose to add a secondary \textit{embedding
	loss} $\mathcal{L}_\text{emb}$ for the meta learner alongside the classification loss $\mathcal{L}_\text{cls}$ used for the prediction network. Recall that we adopt a shared binary classifier in $\mathcal{F}$ to predict the
compatibility of the task description and the input image. To be able to distinguish which task (i.e., class) 
the image belong to, instead of using a binary cross-entropy loss directly, we adopt a calibrated multi-class
cross-entropy loss~\cite{yang2018learning} defined as 
\begin{equation}
\mathcal{L}_\text{cls} = - \frac{1}{N}\sum_{i=1}^N\sum_{t=1}^T \log\left[\frac{\exp(\mathcal{F}(\Vec{x}_i;\theta_t))\cdot y^i_t}{\sum_{j=1}^T\exp(\mathcal{F}(\Vec{x}_i;\theta_j))} \right],
\end{equation}
where $x_i$ is the $i$-th sample in the dataset with size $N$ and $\Vec{y}_i\in\{0,1\}^T$ is the one-hot encoding of the ground-truth labels. T is the number of tasks either in the whole dataset or in the minibatch during 
training. 

For the embedding loss, the idea is to project the latent task embedding $\Vec{e}_t = \mathcal{T}(\Vec{t})$ into a
joint embedding space with the task-aware feature embedding (TAFE). We adopt a metric learning approach that for positive inputs of a
given task, the corresponding TAFE is 
closer to the task embedding $\Vec{e}_{\Vec{t}}$ while for negative inputs, the corresponding TAFE is
far from the task embedding as illustrated in Figure~\ref{fig:tafe}. We use a hinged cosine similarity as the distance measurement  
(i.e. $\phi(p, q) = \max(\texttt{cosine\_sim}(p, q), 0)$)  and  the embedding loss is defined as 
\begin{equation}
\mathcal{L}_\text{emb} =\frac{1}{NT}\sum_i^N \sum_t^T{||\phi(\text{TAFE}(\Vec{x}_i; \theta_t), \Vec{e}_t)- y_t^i||_2^2}.
\end{equation}
We find in experiments this additional supervision helps training the meta learner especially under the case where the number of training tasks is extremely limited. So far, we can define the overall objective as 
\begin{equation}
\min_{\theta, \eta} \mathcal{L} = \min_{\theta, \eta} \mathcal{L}_\text{cls}
+ \beta \cdot \mathcal{L}_\text{emb},
\end{equation}
where $\beta$ is the hyper-parameter to balance the two terms. We use $\beta$
as 0.1 in our experiments if not specified.

	\subsection{Applications}
\label{sec:app}
We now describe how \model design can be utilized in various applications (e.g., zero-shot learning, 
unseen attribute-object recognition and few shot learning) and specify the task descriptions adopted in this 
work. 

\minisection{Zero-shot learning.} In the zero-shot learning (ZSL) setting, the set of classes seen during 
training and evaluated during testing are disjoint~\cite{lampert2014attribute,akata2016label}. Specifically,
let the training set be $\mathcal{D}_s = \{(x, t, y)|x\in\mathcal{X}, t\in\mathcal{T}, y\in\mathcal{Y}\}$, and 
the testing set be $\mathcal{D}_u = \{(x, u, z)|x\in\mathcal{X}, u\in\mathcal{U}, z\in\mathcal{Z}\}$, where $\mathcal{T}\cap\mathcal{U}=\phi$, $|T|=|Y|$ and $|U|=|Z|$.  In benchmark datasets (e.g., CUB~\cite{WelinderEtal2010}, AWA~\cite{lampert2009learning}), each image category 
is associated with an attribute vector, which can be used as the task description in our work.
The goal is to learn a classifier $f_\text{zsl}: \mathcal{X}\rightarrow \mathcal{Z}$. More recently,
Xian~\etal~\cite{xian2018zero} proposed the generalized zero-shot learning (GZSL) setting which is more realistic compared to ZSL. The GZSL setting involves classifying test examples from
both seen and unseen classes, with no prior distinction between them. The classifier in GZSL maps $\mathcal{X}$
to $\mathcal{Y}\cup\mathcal{Z}$. We consider both the ZSL and GZSL settings in our work. 

\minisection{Unseen attribute-object pair recognition.} Motivated by the human capability to compose and recognize
novel visual concepts, Misra \etal~\cite{misra2017red} recently 
proposed a new recognition task to predict unseen compositions of a given set of attributes (e.g., red, modern, ancient, etc) and objects (e.g., banana, city, car, etc) during 
testing and only a subset of attribute-object pairs are seen during training. This can be viewed as a zero-shot 
learning problem but requires more understanding of the contextuality of the attributes. In our work, the 
attribute-object pairs are used as the task descriptions. 

\minisection{Few-shot Learning.} 
In few-shot learning, there are one or a few examples from the novel classes and plenty of examples in the base classes~\cite{hariharan2017low}. 
The goal is to learn a classifier that can classify examples from both the novel and base classes. The sample
image features from different categories can be used as the task descriptions for \modelplural.

	\newcommand{\controlmodel}{controller\xspace}
\newcommand{\controlmodelplural}{controllers\xspace}
\section{Experiments}
\label{sec:exp}
We evaluate our \modelplural on three tasks: zero-shot learning (Section~\ref{sec:zeroshot}), unseen
attribute-object composition (Section~\ref{sec:concept} and few-shot learning (Section~\ref{sec:fewshot}). We observe that TAFE-Net is highly effective in generalizing to new tasks or concepts and is able to match or exceed the state-of-the-art on all the tasks. 

\minisection{Model configurations.}
We first describe the network configurations. The task embedding network $\mathcal{T}$ is a 3-layer FC network with the hidden unit size of 2048 except 
for the aPY dataset~\cite{farhadi2009describing} where we choose  $\mathcal{T}$ as a 2-layer FC network with the hidden size of 2048 to avoid overfitting. The weight generator $
\Vec{g}^i$ is a single FC layer with the output dimension matching the output dimension
of the corresponding feature layer in $\mathcal{F}$. For the prediction network $\mathcal{F}$, 
the TAFE is generated through a 3-layer FC network with the hidden size of 2048 with input image features 
extracted from different pre-trained backbones (e.g., ResNet-18, ResNet-101, VGG-16, VGG-19, etc.)


\subsection{Zero-shot Learning}
\label{sec:zeroshot}
\minisection{Datasets and evaluation metrics}. We conduct our experiments on 5 benchmark datasets:
SUN~\cite{xiao2010sun}, CUB~\cite{welinder2010caltech}, AWA1~\cite{lampert2009learning}, AWA2~\cite{xian2018zero} and aPY ~\cite{farhadi2009describing}, which have different numbers of categories and granularity. In particular, there are only 20 classes (i.e. tasks) available in the aPY dataset while 645 classes are available for training in the SUN dataset. The dataset statistics are shown in Table ~\ref{tab:datasets}.

Following the settings proposed by Xian~\etal, we consider both the generalized zero-shot learning (GZSL) and
the conventional zero-shot learning (ZSL). For GZSL, we report the average per class top-1 accuracy of both unseen $acc_u$ and seen classes $acc_s$ and the harmonic mean $H = 2 \times (acc_u \times acc_s)/(acc_u+acc_s)$. For conventional ZSL, we report the average per-class top-1 accuracy of the unseen classes and adopt the new split provided by Xian~\etal~\cite{xian2018zero}.

\minisection{Training details}. We set the batch size to 32 and use Adam~\cite{kingma2014adam} as
the optimizer with the initial learning rate of $10^{-4}$ for the prediction network and weight generators, and $10^{-5}$ for the task embedding network. 
We reduce the learning rate by 10$\times$ at epoch 30 and 45, and
train the network for 60 epochs. For AWA1, we train the network for 10 epochs and reduce the learning rate by 10$\times$ at epoch 5. 

\minisection{Baselines.} We compare our model with two lines of prior works in our experiments. (1) Discriminative baselines which focus on mapping the images into a rich semantic embedding space. We include the recent 
competitive baselines: LATEM~\cite{zhang2016zero}, ALE~\cite{akata2016label},  DeViSE~\cite{frome2013devise}, 
SJE~\cite{akata2015evaluation}, SYNC~\cite{Changpinyo_2016_CVPR}, DEM~\cite{zhang2017learning} and the newly proposed RelationNet~\cite{yang2018learning}. (2) Generative models that tackle the data
scarcity problem by generating synthetic images for the unseen classes using a GAN~\cite{goodfellow2014generative, zhu2017unpaired} based approach. The generative models can combine different
discriminative models as base networks~\cite{xian2018feature,wang2018low}.
We conduct comparison with f-CLSWGAN~\cite{xian2018feature}, SE~\cite{verma2017generalized}, SP-AEN~\cite{chen2018zero} in this category. Our model falls into the discriminative model category requiring no additional 
synthetic data.

\begin{table}[t]
	\small
	\centering
	\caption{\small Datasets used in GZSL}
	\addtolength{\tabcolsep}{-1pt}
	\label{tab:datasets}
	\adjustbox{width=.9\linewidth}{
		\begin{tabular}{@{}cccccc@{}}
			Dataset            & SUN   & CUB   & AWA1   & AWA2   & aPY    \\ \midrule
			No. of Images      & 14,340 & 11,788 & 30,475  & 37,322  & 15,339  \\
			Attributes Dim.  & 102   & 312   & 85     & 85     & 64     \\
			$\mathcal{Y}$      & 717   & 200   & 50     & 50     & 32     \\
			$\mathcal{Y}^{seen}$ & 645   & 150   & 40     & 40     & 20     \\
			$\mathcal{Y}^{unseen}$ & 72    & 50    & 10     & 10     & 12     \\
			Granularity        & fine  & fine  & coarse & coarse & coarse \\ \bottomrule
	\end{tabular}}
\end{table}

\begin{table*}[t]
	\small
	\addtolength{\tabcolsep}{-1pt}
	\centering
	\caption{\small Evaluate \model on five standard benchmarks under the ZSL and the GZSL settings. Models with $^\dagger$ (f-CLSWGAN, SE and SP-AEC) generate additional data for training while the remaining models do not.  \textbf{\textcolor{red}{Red}} denotes the best performing model on each dataset
		and \textbf{\textcolor{blue}{blue}} denotes the prior art of discriminative models. Our model is better than all the other discriminative models and also competitive compared to models with additional synthetic data.}
	\label{tab:zs_1}
	\adjustbox{width=\linewidth}{
		\begin{tabular}{@{}l|ccccc|ccc|ccc|ccc|ccc|ccc@{}}
			\multirow{3}{*}{Method} & \multicolumn{5}{c|}{Zero-shot Learning} & \multicolumn{15}{c}{Generalized Zero-shot Learning} \\ 
			&SUN & CUB & AWA1 & AWA2 & aPY & \multicolumn{3}{c|}{SUN} & \multicolumn{3}{c|}{CUB} & \multicolumn{3}{c|}{AWA1} & \multicolumn{3}{c|}{AWA2} & \multicolumn{3}{c}{aPY} \\
			& T1 & T1 & T1 & T1 & T1 & u  & s  & H     & u   & s  & H     & u   & s   & H  & u   & s   & H     & u   & s     & H  \\ \midrule
			LATEM~\cite{zhang2016zero} & 55.3 & 49.3 &  55.1 & 55.8 & 35.2   & 14.7    & 28.8  & 19.5  & 15.2     & 57.3  & 24.0  & 7.3      & 71.7   & 13.3  & 11.5     & 77.3   & 20.0  & 0.1 & 73.0  & 0.2   \\
			ALE~\cite{akata2016label}  & 58.1 & 54.9 & 59.9 & 62.5 & 39.7   & 21.8    & 33.1  & \textbf{\textcolor{blue}{26.3}}  & 23.7     & 62.8  & 34.4  & 16.8     & 76.1   & 27.5  & 14.0     & 81.8   & 23.9  & 4.6      & 73.7  & 8.7   \\
			DeViSE\cite{frome2013devise}  & 56.5 &  52 & 54.2 & 59.7 & \textbf{\textcolor{blue}{39.8}} & 16.9    & 27.4  & 20.9  & 23.8     & 53.0  & 32.8  & 13.4     & 68.7   & 22.4  & 17.1     & 74.7   & 27.8  & 4.9      & 76.9  & 9.2   \\
			SJE~\cite{akata2015evaluation} & 53.7 & 53.9 & 65.6 & 61.9 & 32.9 & 14.7    & 80.5  & 19.8  & 23.5     & 59.2  & 33.6  & 11.3     & 74.6   & 19.6  & 8.0      & 73.9   & 14.4  & 3.7      & 55.7  & 6.9   \\
			ESZSL~\cite{romera2015embarrassingly} &    54.5 & 53.9 & 58.2 & 58.6 & 38.3  & 11.0    & 27.9  & 15.8  & 12.6     & 63.8  & 21.0  & 6.6      & 75.6   & 12.1  & 5.9      & 77.8   & 11.0  & 2.4      & 70.1  & 4.6   \\
			SYNC ~\cite{Changpinyo_2016_CVPR}  & 56.3 & \textbf{\textcolor{blue}{55.6}} & 54.0 & 46.6 & 23.9 & 7.9     & 43.3  & 13.4  & 11.5     & 70.9  & 19.8  & 8.9      & 87.3   & 16.2  & 10.0     & 90.5   & 18.0  & 7.4      & 66.3  & 13.3  \\
			RelationNet~\cite{yang2018learning} &- & \textbf{\textcolor{blue}{55.6}} & 68.2 & 64.2 & - & - & - & - & 38.1 & 61.1 & \textbf{\textcolor{blue}{47.0}} & 31.4 & 91.3 & 46.7 & 30.0 & 93.4 & \textbf{\textcolor{blue}{45.3}} & - & - & - \\
			DEM~\cite{zhang2017learning} & \textbf{\textcolor{blue}{61.9}} & 51.7 & \textbf{\textcolor{blue}{68.4}} & \textbf{\textcolor{blue}{67.1}} & 35.0 & 20.5 & 34.3 & 25.6 & 19.6 & 57.9 & 29.2 & 32.8 & 84.7 & \textbf{\textcolor{blue}{47.3}} & 30.5 & 86.4 & 45.1 & 11.1 & 75.1 & \textbf{\textcolor{blue}{19.4}} \\
			\midrule
			f-CLSWGAN$^\dagger$~\cite{xian2018feature} & 60.8 & 57.3 & 68.2 & - & - &42.6&36.6& \textbf{\textcolor{red}{39.4}}& 57.7 & 43.7 & \textbf{\textcolor{red}{49.7}}&
			61.4 & 57.9 & 59.6 &-&-&-& -&-&- \\
			SE$^\dagger$~\cite{verma2017generalized} & \textbf{\textcolor{red}{63.4}} & \textbf{\textcolor{red}{59.6}} & 69.5 & 69.2 & - & 40.9 & 30.5 & 34.9 & 53.3 & 41.5 & 46.7 
			& 67.8 & 56.3 & 61.5  & 58.3 & 68.1& \textbf{\textcolor{red}{62.8}}& -&-&- \\
			SP-AEN$^\dagger$~\cite{chen2018zero} & 59.2 & 55.4 & - & 58.5 & 24.1  & 24.9 & 38.6 & 30.3 & 34.7 & 70.6 & 46.6 & - & - & - & 23.3 & 90.9 & 37.1 & 13.7 & 63.4 & 22.6 \\
			\midrule
			\model & 60.9 & 56.9 & \textbf{\textcolor{red}{70.8}} & \textbf{\textcolor{red}{69.3}} & \textbf{\textcolor{red}{42.2}} & 27.9 & 40.2 & 33.0 & 41.0 & 61.4 & 49.2 & 50.5 & 84.4 & \textbf{\textcolor{red}{63.2}} & 36.7 & 90.6 &52.2 & 24.3 & 75.4 & \textbf{\textcolor{red}{36.8}}\\
			\bottomrule
	\end{tabular}}
\end{table*}

\minisection{Quantitative results}. We compare the performance of \model to the prior works in Table~\ref{tab:zs_1}. Overall, our model outperforms 
existing approaches including the generative models on the AWA1, AWA2
and aPY datasets under the ZSL setting and on the AWA1 and aPY datasets 
under the GZSL setting.
\model outperforms the discriminative models (denoted in \textbf{\textcolor{blue}{blue}} in Table~\ref{tab:zs_1}) by a large margin (e.g., roughly 16 points improvement on AWA1 and 17 points on aPY) on the GZSL test. For the more challenging fine-grained SUN and CUB datasets, we are able to improve the results by 7 and 2 points. The results indicate that better embeddings can aid in model generalization.



\begin{table}[t]
	\small
	\centering
	\caption{\small Ablation of the embedding loss on the five benchmarks under GZSL. Harmonic mean (H) is reported.}
	\label{tab:zs_ab}
	\adjustbox{width=\linewidth}{
		\begin{tabular}{@{}lc|c|c|c|c@{}}
			Method & SUN & CUB & AWA1 & AWA2 & aPY \\
			\midrule
			\model w/o EmbLoss & \textbf{33.1} & 45.4 & 58.8 & 47.2 & 30.5 \\
			\model & 33.0 & \textbf{49.2} & \textbf{{63.2}} & \textbf{52.2} & \textbf{{36.8}}\\
			\bottomrule
	\end{tabular}}
	\vspace{-1em}
\end{table}

\minisection{Embedding loss ablation}. 
We provide the harmonic mean of our models with and without the embedding loss under the GZSL setting on five benchmark datasets in
Table~\ref{tab:zs_ab}. 
In general, models with the embedding loss outperform those 
without the embedding loss except for the SUN dataset whose number of
categories is about $3$ to $22\times$ larger than the other datasets.  This observation matches our assumption
that the additional supervision on the joint embedding better addresses the data scarcity (i.e. fewer
class descriptions than the visual inputs) of training the controller model. 

\minisection{Embedding visualization.}
\begin{figure}[t]
	\centering
	\begin{subfigure}[t]{.8\linewidth}
		\centering
		\includegraphics[width=\linewidth]{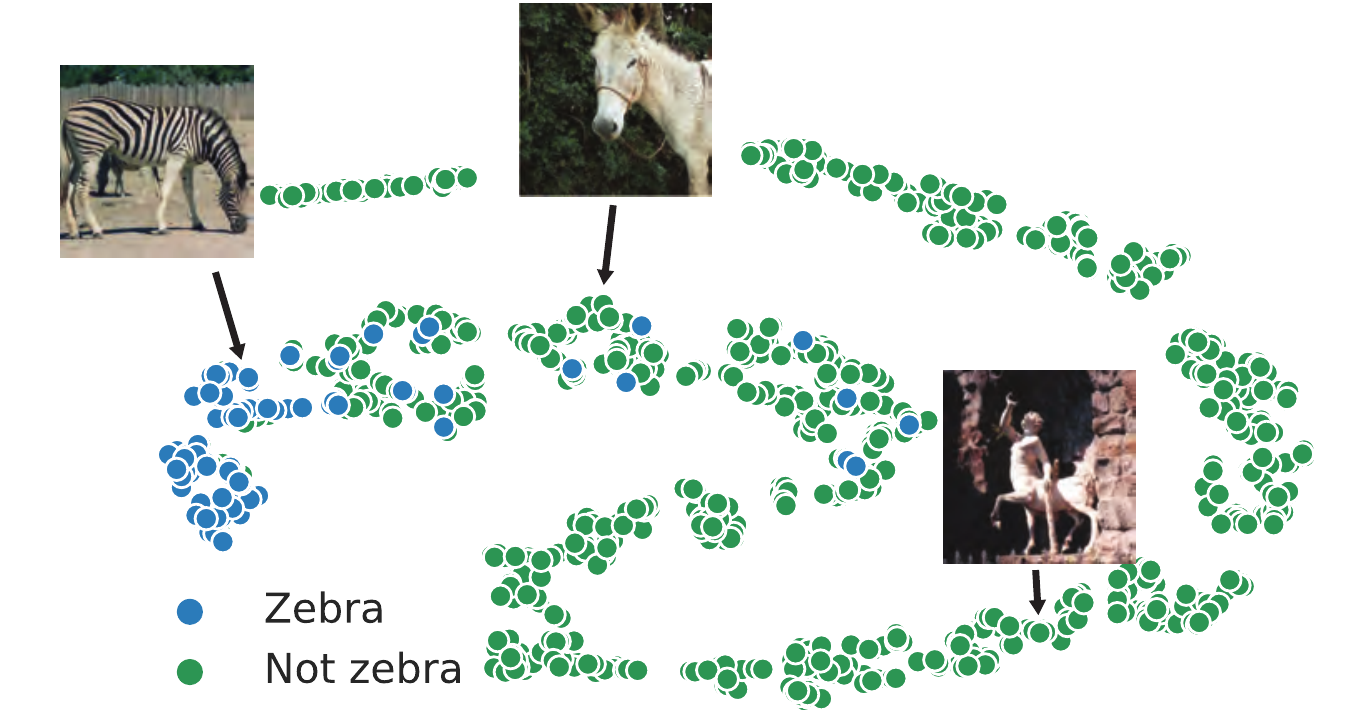}
	\end{subfigure}%
	\\
	\vspace{2em}
	\begin{subfigure}[t]{.8\linewidth}
		\centering
		\includegraphics[width=\linewidth]{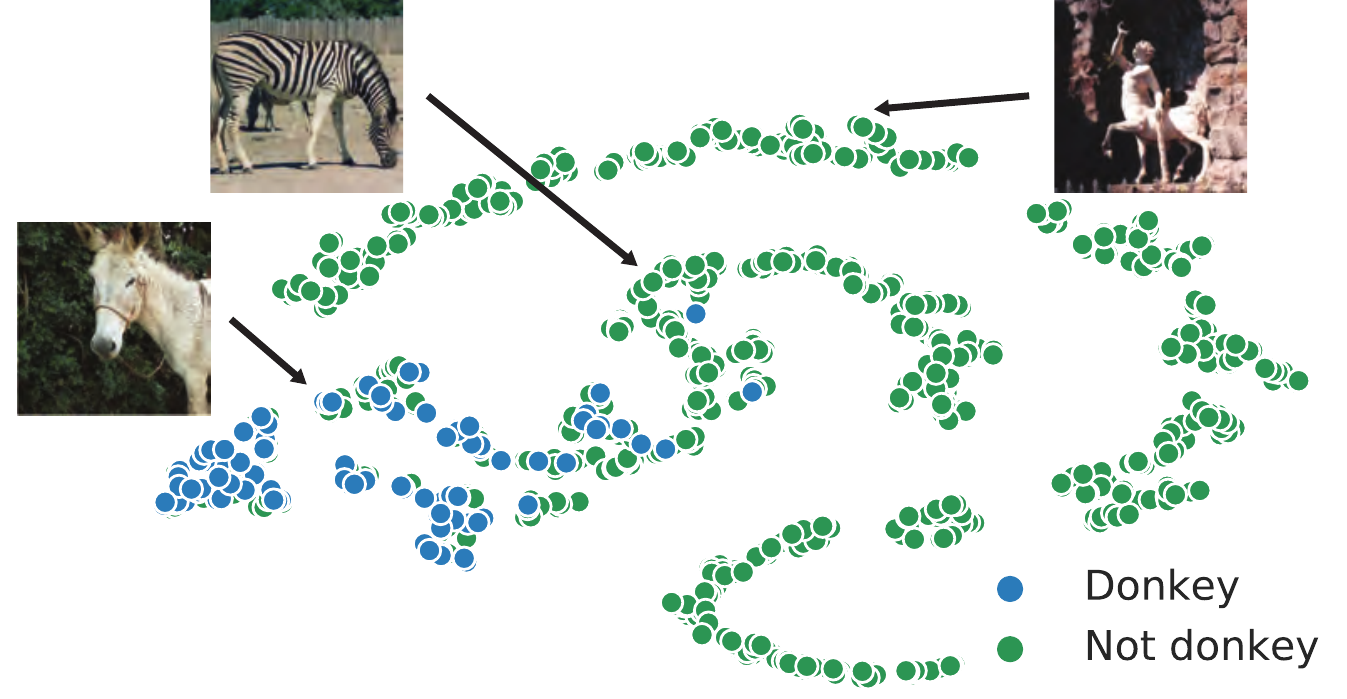}
	\end{subfigure}
	\caption{Task-aware Image Feature Embedding projected into two dimensions using t-SNE~\cite{tsne} for two tasks (Zebra and Donkey). Note that changing the task produces different embeddings for the same data.\vspace{-1em}}
	\label{fig:tafe-vis}
\end{figure}
In Figure~\ref{fig:tafe-vis}, we visualize the task-aware feature embeddings of images from the aPY dataset under different task descriptions. As we can see, image embeddings of the same image are projected into different clusters conditioned on the task descriptions. 

\subsection{Unseen Visual-attribute Composition}
\label{sec:concept}
Besides the standard zero-shot learning benchmarks, we evaluate our model on the visual-attribute composition task proposed by Misra~\etal~\cite{misra2017red}. The goal is to compose a set of visual concept primitives like attributes 
and objects (e.g. \texttt{large elephant, old building,} etc.) to obtain new visual concepts for a given image. This is a more challenging ``zero-shot'' learning task,  which requires the model not only to predict unseen visual concept compositions but also to model the contextuality of the concepts.

\begin{figure}[t]
	\centering
	\setlength{\tabcolsep}{0.2em}
	\adjustbox{width=\linewidth}{
		\begin{tabular}{ccc}
			\includegraphics[width=0.4\linewidth]{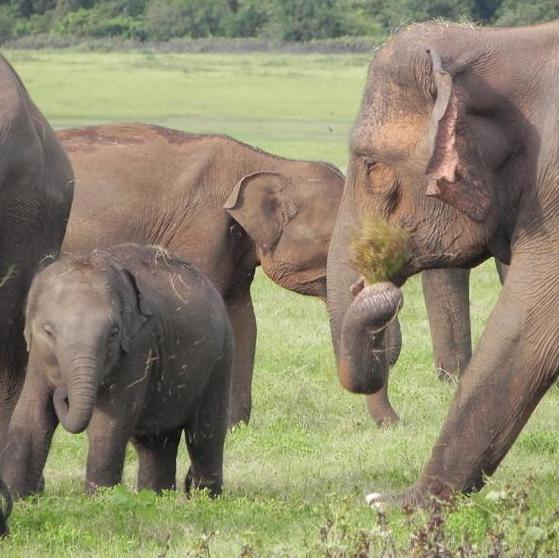} & 
			\includegraphics[width=0.4\linewidth]{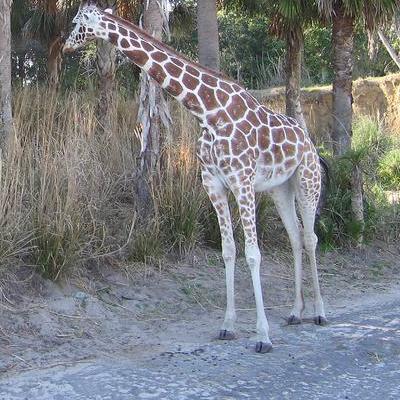} &
			\includegraphics[width=0.4\linewidth]{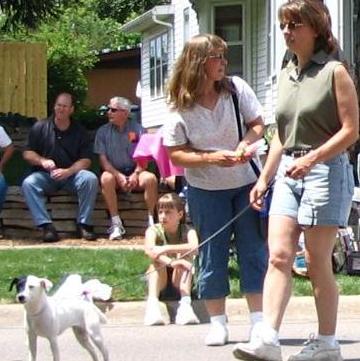} \\
			\includegraphics[width=0.4\linewidth]{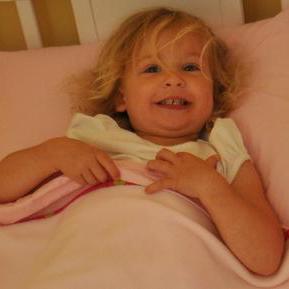} &
			\includegraphics[width=0.4\linewidth]{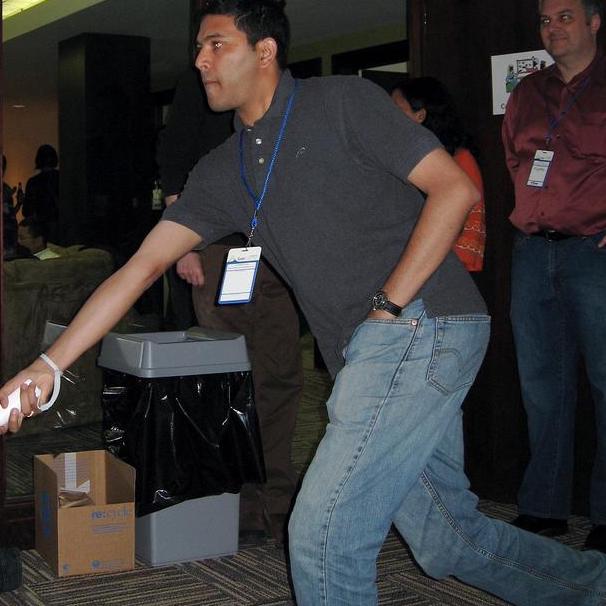} &
			\includegraphics[width=0.4\linewidth]{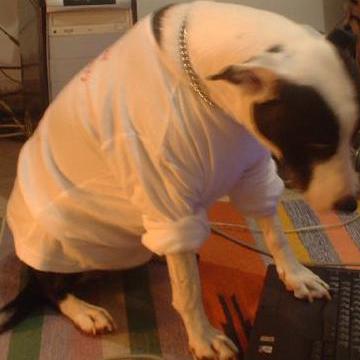} \\
	\end{tabular}}
	\caption{\small Samples in StanfordVRD. Each image 
		is described by a Subject-Verb-Object triplet. From top left to the bottom right: (elephant, on, grass), (giraffe, in, street), (person, walk, dog), 
		(pillow, behind, person), (person, wears, jeans), (dog, has, shirt).}
	\label{fig:stanfordvrd_imgs}
\end{figure}

\begin{figure*}[t!]
	\centering
	\small
	\adjustbox{width=\linewidth}{
		\begin{tabular}{c | c}
			Modern City & Ancient Town \\
			\includegraphics[width=0.49\linewidth]{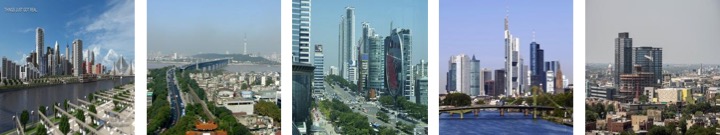}
			&
			\includegraphics[width=0.49\linewidth]{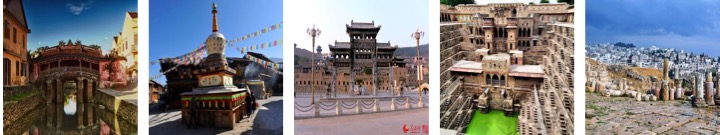}
			\\
			Modern Clock & Ancient Clock \\
			\includegraphics[width=0.49\linewidth]{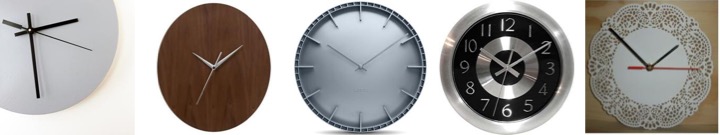}
			&
			\includegraphics[width=0.49\linewidth]{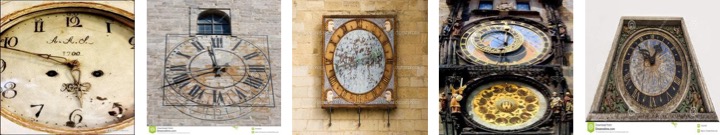}
			\\
			Sunny Valley & Sunny Creek \\
			\includegraphics[width=0.49\linewidth]{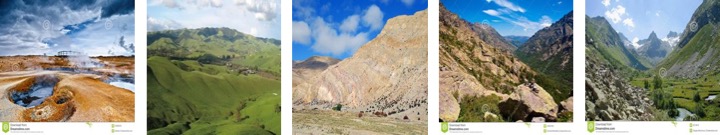}
			&
			\includegraphics[width=0.49\linewidth]{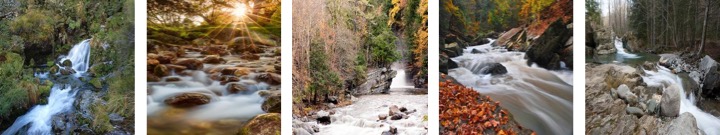}
			\\
	\end{tabular}}
	\caption{Top retrievals on the unseen pairs of the MITStates dataset. Our model can learn to compose new concepts from the existing attributes and objects
		while respecting their context. The second row shows some of the failure cases.}
	\label{fig:retrieval}
\end{figure*}

\minisection{Datasets and evaluation metrics}. We conduct the experiments on two
datasets: MITStates~\cite{StatesAndTransformations} (image samples in Figure~\ref{fig:retrieval}) and
the modified StanfordVRD~\cite{lu2016visual} (image samples in Figure~\ref{fig:stanfordvrd_imgs}). The setup is the same as Misra~\etal~\cite{misra2017red}.
Each image in the MITStates dataset is assigned a pair of (attribute, object) as
its label. The model is trained on 34K images with 1,292 label pairs and tested on
19K images with 700 unseen pairs. The second dataset is constructed based on 
the bounding box annotations of the StanfordVRD dataset. Each sample has an SPO (subject, predicate, object)
tuple as the ground truth label.  The dataset has 7,701 SPO triplets and 1,029 of them
are seen only in the test split. We evaluate our models only on examples with unseen
labels. We extract the image features with
pre-trained models on ImageNet. We use VGG-16 and ResNet-101 as our main
feature extractors and also test features extracted 
with VGG-19 and ResNet-18 for ablation. 
For the task descriptions, we concatenate the word embeddings of the attributes
and objects with word2vec~\cite{mikolov2013efficient} trained with GoogleNews. We also consider one-hot encoding for the task ID in the ablation. 

For evaluation metrics, we report the mean Average Precision (mAP) of images 
with unseen labels in the test set together with the top-$k$ accuracy where
$k=1,2,3$. We follow the same training schedule as that used in the zero shot learning experiments.

\begin{table}[t]
	\centering
	\caption{\small Evaluation on 700 unseen (attribute, object) pairs on 19K images of the MITStates Dataset and 1029 unseen SPO triplets on 1000 images of the StanfordVRD Dataset. \model improves over the baselines by a large margin.}
	\label{tab:mit}
	\small
	\adjustbox{width=\linewidth}{
		\begin{tabular}{lcccc|cccc}
			&  \multicolumn{4}{c}{MITStates} & \multicolumn{4}{c}{StanfordVRD} \\
			\multirow{2}{*}{Method} & \multirow{2}{*}{AP} &  \multicolumn{3}{c}{Top-$k$ Accuracy} & \multirow{2}{*}{AP} &  \multicolumn{3}{c}{Top-$k$ Accuracy} \\
			&   & 1 & 2 & 3 & & 1 & 2 & 3 \\ \midrule
			Visual Product~\cite{misra2017red} & 8.8 & 9.8 & 16.1 & 20.6 & 4.9 & 3.2 & 5.6 & 7.6 \\
			Label Embed (LE)~\cite{misra2017red} & 7.9 & 11.2  & 17.6 & 
			22.4 &  4.3 & 4.1  & 7.2  & 10.6  \\
			LEOR~\cite{misra2017red} & 4.1 & 4.5  & 6.2 & 11.8 & 
			0.9 & 1.1 & 1.3 & 1.3 \\
			LE + R~\cite{misra2017red} & 6.7 & 9.3 & 16.3 & 20.8 &
			3.9 & 3.9 & 7.1 & 10.4 \\
			Red Wine~\cite{misra2017red} & 10.4  & 13.1 & 21.2 & 27.6 &
			5.7  & 6.3 & 9.2  & 12.7 \\ \midrule
			\model& \textbf{16.3} & \textbf{16.4} & \textbf{26.4} & \textbf{33.0} & \textbf{12.2}  & \textbf{12.3}  &\textbf{19.7} & \textbf{27.5} \\
			\bottomrule
	\end{tabular}}
	\vspace{-1em}
\end{table}

\minisection{Quantitative results.} We compare our model with
several baselines provided by Misra~\etal~\cite{misra2017red} and
summarize the results in
Table~\ref{tab:mit} on both the MITStates and StanfordVRD
datasets. Our model surpasses the state-of-the-art models
with an improvement
of more than 6 points in mAP and 4 to 15 points in top-$k$
accuracy. Nagarajan and Grauman~\cite{nagarajan2018attrop}
recently proposed an embedding learning framework for visual-attribute composition. They report the top-1 
accuracy of $12.0\%$ on the MITStates dataset with
ResNet-18 features. For fair
comparison, we use the same ResNet-18 features and obtain
the top-1 accuracy of $15.1\%$. 

\begin{table}[t]
	\centering
	\small
	\caption{\small Ablation study with different task encoding and base network features. The variance of performance of \model under different settings is minimal.}
	\label{tab:comp_ab}
	\adjustbox{width=.85\linewidth}{
		\renewcommand{\arraystretch}{1.27}
		\begin{tabular}{lccccc}
			\multirow{2}{*}{Task Encoding} & \multirow{2}{*}{Features}  & \multirow{2}{*}{AP} & \multicolumn{3}{c}{Top-$k$ Accuracy} \\
			&           &          & 1          & 2          & 3          \\ \midrule
			Word2vec  & ResNet-101 & 16.2 & 17.2    & 27.8       & 35.7       \\
			Onehot    & ResNet-101 & 16.1  & 16.1   & 26.8       & 33.8       \\
			\midrule
			Word2vec  & VGG16  & 16.3  & 16.4       & 26.4       & 33.0       \\
			Onehot    & VGG16  & 16.3  & 16.4       & 25.9       & 32.5       \\
			\midrule
			Word2vec  & VGG19  & 15.6  & 16.2       & 26.0       & 32.4       \\
			Onehot    & VGG19  & 16.3  & 16.4       & 26.0       & 33.1       \\
			\bottomrule
	\end{tabular}}
\end{table}

\minisection{Ablation on the feature extractor and task description}. We consider 
different feature extractors (ResNet-101, VGG-16 and 19) and task encodings (word2vec and one-hot encoding) for ablation and summarize the results in Table~\ref{tab:comp_ab}. The average precision difference between different feature
extractors are very minimal (within $0.1\%$) and the largest gap in Top-3 accuracy 
is within $2\%$. This indicates that \model is robust in transforming the 
generic features into task-aware feature embeddings. For the task encoding,
the one-hot encoding is comparable to the word2vec encoding and even stronger 
when using VGG-19 features. This shows that the task transformer network $\mathcal{T}$ is very expressive to extract rich semantic information simply from the
task IDs.

\minisection{Visualization}. 
In Figure~\ref{fig:retrieval}, we show the top
retrievals of unseen attribute-object pairs from the MITStates dataset. 
Our model can learn to compose new concepts from the existing attributes and objects
while respecting their context.


\subsection{Few-shot Image Classification}
\label{sec:fewshot}
Our model naturally fits the few-shot learning setting where one or few images
of a certain category are used as the task descriptions. Unlike prior work on meta-learning 
which experiments with few classes and low resolution images~\cite{vinyals2016matching, snell2017prototypical,finn2017model}, we evaluate
our model on the challenging benchmark proposed by Hariharan and Girshick~\cite{hariharan2017low}. The benchmark is based on the ImageNet images and
contains hundreds of classes that are divided into base classes and novel classes.
At inference time, the model is provided with one or a few examples from the novel classes
and hundreds of examples from the base classes. 
The goal is to obtain high accuracy 
on the novel classes without sacrificing the performance on the base classes. 

\minisection{Baselines}. In our experiments, the  baselines we consider are the 
state-of-the-art meta learning models: Matching Network (MN)~\cite{vinyals2016matching}
and Prototypical Network (PN)~\cite{snell2017prototypical}. We also compare the logistic regression (LogReg) baseline provided by Hariharan and Girshick~\cite{hariharan2017low}. Another line of research~\cite{wang2018low, hariharan2017low} for few-shot learning is to combine
the meta-learner with a ``hallucinator'' to generate additional training data. We
regard these works as complementary approaches to our meta-learning model. 

\begin{table}[t]
	\centering
	\small
	\caption{Few-shot ImageNet Classification on ImageNet. Our model
		is competitive compared to the state-of-the-art meta learning model
		without hallucinator.}
	\label{tab:few-shot}
	\begin{tabular}{@{}l|cc|cc@{}}
		Method & \multicolumn{2}{c}{Novel Top-5 Acc} & \multicolumn{2}{|c}{All Top-5 Acc} \\
		& n=1 & n=2  & n=1 & n=2  \\ 
		\midrule
		LogReg ~\cite{hariharan2017low} & 38.4 & 51.1  & 40.8 & 49.9 \\
		PN~\cite{snell2017prototypical} & 39.3 & 54.4  &  49.5 & 61.0 \\
		MN~\cite{vinyals2016matching} &  43.6 & 54.0  &  54.4 & 61.0  \\
		\midrule
		\model & 43.0 & 53.9 & \textbf{55.7} & \textbf{61.9} \\
		\midrule\midrule
		LogReg w/ Analogies~\cite{hariharan2017low} & 40.7 & 50.8 & 52.2 & 59.4 \\
		PN w/ G ~\cite{wang2018low} & 45.0 & 55.9 & 56.9 & 63.2 \\ 
		\bottomrule
	\end{tabular}%
	\vspace{-1em}
\end{table}

\minisection{Experiment details}. We follow the prior works~\cite{hariharan2017low, wang2018low} to run five trials for each setting of n (the number of examples per
novel class, n = 1 and 2 in our experiments) on the five different data splits and report the average top-5 accuracy of both
the novel and all classes. We use the features trained with ResNet-10 using SGM
loss provided by Hariharan and Girshick~\cite{hariharan2017low} as inputs. 
For training, we sample 100
classes in each iteration and use SGD with momentum of 0.9 as the optimizer. The
initial learning rate is set to 0.1 except for the task embedding network (set to
0.01) and the learning rate is reduced by 10$\times$ every 8k iterations.
The model is trained for 
30k iterations in total. Other hyper-paramters are set to the same as Hariharan and Girshick~\cite{hariharan2017low} if not mentioned.

\minisection{Quantitative results}. As shown in Table~\ref{tab:few-shot}, our
model is on par with  state-of-the-art meta learning models on the 
novel classes while outperforming them on all categories. Attaching a ``hallucinator'' to the meta learning model improves performance in
general. Our model can be easily attached with a hallucinator and we leave
the detailed study as future work due to the time constraint.

	\section{Conclusion}
In this work, we explored a meta learning based approach to generate task aware feature embeddings for settings with little or no training data.
We proposed TAFE-Net, a network that generates task aware feature embeddings (TAFE) conditioned on the given task descriptions.
\model has a task-aware 
meta learner that generates weights for the feature embedding layers in a standard prediction network.
To address the challenges in training the meta learner, we introduced two key
innovations: (1) adding an additional embedding loss to improve the generalization of the meta learner; (2) a novel weight factorization
scheme to generate parameters of the prediction network more effectively. 
We demonstrated the general applicability of the proposed network design on a range of benchmarks in  zero-/few- shot learning,  and matched or exceeded the state-of-the-art.

	\vspace{-1em}
	\subsubsection*{Acknowledgments}
	This work was supported by Berkeley AI Research, RISE Lab and Berkeley DeepDrive. In addition to NSF CISE Expeditions Award CCF-1730628, this research is supported by gifts from Alibaba,
	Amazon Web Services, Ant Financial, Arm, CapitalOne,
	Ericsson, Facebook, Google, Huawei, Intel, Microsoft,
	Nvidia, Scotiabank, Splunk and VMware.
	
	{\small
		\bibliographystyle{ieee}
		\bibliography{references}
	}

\end{document}


\title{TAFE-Net: Task-Aware Feature Embeddings for \\ Efficient Learning and Inference \\
\textcolor{red}{\large Supplementary Material}}

\author{First Author\\
Institution1\\
Institution1 address\\
{\tt\small firstauthor@i1.org}
\and
Second Author\\
Institution2\\
First line of institution2 address\\
{\tt\small secondauthor@i2.org}
}

\maketitle

\section*{Appendix}
In the appendix, we are going to investigate that how critical the proposed task-aware feature embedding is for the improved
performance of \model as it adjusts the image feature based on the task 
specification. We present more analysis over the learned embeddings (task-aware feature embeddings and latent task embedding of the meta learner) in \model. Moreover, 
we show experimental results  and visualization on the latent task embedding to demonstrate that the 
embedding network in the task-aware meta learner is able to capture the visual/semantic
information and thus embeddings of semantically similar tasks are closer 
to those of semantically different tasks even when the task encoding itself does 
not contain any semantic information (e.g., one-hot encoding). 


\section*{A.1 Task-aware Embedding Visualization}
\label{sec:tafe}
We now provide more visualization for the task-aware feature embedding (TAFE). 
The main characteristics of TAFE is the same image input may have different feature 
embeddings under different tasks/contexts. For example, consider the task: 
\textit{is this a building or statue?} which can be further decomposed to binary tasks:
\textit{is this a building} and \textit{is this a statue?} The image of a 
building
will be projected to a \texttt{building} cluster and a \texttt{non-statue} cluster. 

To this end, we visualize the trained \model on the aPY dataset under different 
binary task descriptions. We use t-SNE~\cite{tsne} to visualize the TAFEs for 
all test images in the aPY dataset. The blue points stand for the positive examples
of the task and the green points for the negative examples of the task. As shown in Figure~\ref{fig:tsne}, we randomly select three pairs of tasks and demonstrate that 
the same image can be projected to different clusters under different task 
specifications. The learned task-aware image feature is adjusted to the task specification. 

\section*{A.2 Image Shallow Embeddings as Task Desc.}
\label{sec:efficient}
As presented in Section 4.4 in the main paper, the input image itself can be used
as the task description. In this section, we conduct an additional experiment to
analyze the behavior of the meta learner. 
 
\begin{figure*}[h!]
\begin{tabular}{cc}
\includegraphics[width=0.45\textwidth]{figs/tsne_supp/building.pdf} &
\includegraphics[width=0.45\textwidth]{figs/tsne_supp/statue.pdf} \\
\includegraphics[width=0.45\textwidth]{figs/tsne_supp/carriage.pdf} &
\includegraphics[width=0.45\textwidth]{figs/tsne_supp/jetski.pdf} \\
\includegraphics[width=0.45\textwidth]{figs/tsne_supp/monkey.pdf} &
\includegraphics[width=0.45\textwidth]{figs/tsne_supp/goat.pdf} \\
\end{tabular}
\caption{Task-aware Image Feature Embedding projected into two dimensions using t-SNE\cite{tsne} for six different tasks. Note that changing the task produces different embeddings for the same data.}
\label{fig:tsne}
\end{figure*}


\begin{figure*}[h]
    \centering
    \includegraphics[width=\textwidth]{figs/gate_vis.pdf}
    \caption{Weight shuffling. The in-group shuffling has much higher accuracy than the out-of-group shuffling.}
    \label{fig:vis_shuffle}
\end{figure*}

\begin{figure*}[t]
    \centering
    \includegraphics[width=\textwidth]{figs/retrievals_supp.pdf}
    \caption{Top retrievals on the unseen pairs of the MITStates dataset}
    \label{fig:vis_retrievals_supp}
\end{figure*}

\begin{figure*}[h!]
    \centering
    \includegraphics[width=\linewidth]{figs/comp_vis2.png}
    \caption{Task Embedding Visualization with t-SNE on MITState Dataset. The
    figure shows the nearest seen compositions (e.g. \texttt{crushed garlic}, \texttt{duced garlic}) of the unseen composition \texttt{peeled garlic}. }
    \label{fig:comp_vis2}
\end{figure*}

\begin{figure*}[h!]
    \centering
    \includegraphics[width=\linewidth]{figs/comp_vis3.png}
    \caption{Task Embedding Visualization with t-SNE on MITState Dataset. The
    figure shows the nearest seen compositions (e.g. \texttt{broken city}, \texttt{torn city}) of the unseen composition \texttt{shattered city}.}
    \label{fig:comp_vis3}
\end{figure*}

We evaluate the trained \model with VGG-16 used in Section 4.4 as follows: for a given fine-grained class A 
(e.g., dolphin), we re-assign the task embedding for each input in class A with a
randomly chosen task embedding from other classes either within the same coarse category
(referred to as in-group shuffling) or different categories (referred to as out-of-group
shuffling).  In Figure~\ref{fig:vis_shuffle}, we plot the test accuracy of class
\texttt{dolphin}, belonging to the coarse category \texttt{aquatic mammals} with randomly selected task
embeddings (repeated 20 times for each input) from classes in the same coarse category
(in red) and 5 classes for other coarse categories (in blue). The test accuracy with
in-group shuffling is much higher than that that with out-of-group shuffling. Especially, when
applying the task embeddings from the tulip category, the test accuracy drops to 1\%
while the accuracy with in-group shuffling is mostly above 50\%. 

\section*{A.3 Visual Attribute Composition}
\label{sec:composition}
In the visual attribute composition task, the attribute-object pairs are used as 
\textit{task descriptions}. The goal is to predict unseen compositions at test time. In Figure~\ref{fig:vis_retrievals_supp}, we present the top retrievals 
on the unseen attribute-object pairs of the MITStates dataset. 

In our main paper, we use either the word2vec~\cite{mikolov2013efficient} or one-hot encoding to encode
each attribute and object separately and the concatenation is fed into the 
\textit{task-aware meta learner} producing a latent task embedding $\mathcal{T}(t)$. 
The latent task embedding is then used as the input to 
the layer-specific weight generators. 

To study the relation of the unseen and seen compositions, we visualize the latent task embeddings for all compositions with t-SNE~\cite{tsne}. In Figure~\ref{fig:comp_vis2} and Figure~\ref{fig:comp_vis3}, we randomly pick up a 
unseen composition and show its nearest neighbors in the embedding space.  
We find that the unseen compositions are close to the seen compositions that are
semantically related even when one-hot encoding is used as the task encoding.
This implies that through the joint visual-task embedding learning, the task
embedding network is able to capture the visual semantic information effectively.

\clearpage
\clearpage
{\small
\bibliographystyle{ieee}
\bibliography{references}
}